\documentclass[extendedabs]{bmvc2k}
\usepackage[colorlinks = true,
            linkcolor = blue,
            urlcolor  = blue,
            citecolor = blue,
            anchorcolor = blue]{hyperref}

\usepackage{amsmath}

\newcommand{\etal}{\mbox{\emph{et al.\ }}}

\begin{document}

\title{Beyond Semantic Image Segmentation : Exploring Efficient Inference in Video}
\addauthor{
Subarna Tripathi$^1$, 
Serge Belongie$^2$,
Truong Nguyen$^1$
}{}{1}
\addinstitution{
$^1$University of California San Diego. 
$^2$Cornell NYC Tech.
}

\maketitle

%\let\thefootnote\relax\footnote{This is an extended abstract. The full paper is available at the \href{http://www.cv-foundation.org/openaccess/CVPR2015.py}{Computer Vision Foundation webpage}. }
%\vspace{-0.2in}

% Extended abstract begins here.  In a one-page document, there is
% little need for section headers, but you may use \section etc if you
% wish.

\noindent
Deep convolutional neural networks (DCNNs) trained on a large number of images with
pixel-level annotations or a combination of strongly labeled and weakly-labeled images have recently been the state-of-the-art in semantic image segmentation with significant performance improvement. 

However, due to the very invariance properties that make DCNNs good for high level tasks such as classification, visual delineation capacities for deep learning techniques are limited. Recent approaches address this problem with Conditional Random Field (CRF) based graphical model in two ways: \\
\indent
1. Adding a post-processing step of CRF-based probabilistic graphical model for the pixel-level classification \cite{DeepLAB15, PapandreouCMY15}.\\
\indent
2. Integrating the graphical model as a part of the CNN to make the end-to-end learning with the usual back-propagation possible without the need of post-processing \cite{CRFasRCNN15}.\\
\noindent
In either case, the final pixel-level classification accuracy and efficiency remain highly dependent on the inference step of the image-based CRF \cite{DenseCRF13} involved where fast approximate MPM inference is performed using cross bilateral filtering techniques within a mean-field approximation framework.

Alvarez \etal \cite{Alvarez14} demonstrates that performing inference on all test images at once in a dense CRF yields better results than inferring one image at a time without additional computation cost compared to performing segmentation sequentially on individual images. It is to be noted that the dense CRF \cite{DenseCRF13} achieves good results with only unary and pairwise terms. This fully-connected pair-wise model is more expressive than its 4 or 8-connected random field counter-parts. Yet, it lacks the ability to handle high-order terms. Models \cite{Kohli09,Ladicky2009,Ladicky2010} using higher-order terms such as label consistency over large regions (pattern-based potentials) and relations of global co-occurrence potentials, are shown to be more expressive for object class segmentation task. Filter-based inference for those higher-order terms is formulated in \cite{vineet2014filter} which enables significant speed-up compared to those graph-based methods \cite{Kohli09,Ladicky2009,Ladicky2010}. Yet, it needs to consider temporal consistency when applied in co-segmentation or video semantic segmentation.   

We explore the efficiency of the CRF inference module beyond image level semantic segmentation. The key idea is to combine the best of two worlds of semantic co-labeling and exploiting more expressive models. Similar to \cite{Alvarez14} our formulation enables us perform inference over ten thousand images within seconds. On the other hand, it can handle higher-order clique potentials similar to \cite{vineet2014filter} in terms of region-level label consistency and context in terms of co-occurrences. We follow the mean-field updates for higher order potentials similar to \cite{vineet2014filter} and extend the spatial smoothness and appearance kernels \cite{DenseCRF13} to address video data inspired by \cite{Alvarez14}; thus making the system amenable to perform video semantic segmentation most effectively.  

\begin{figure*}[t]
\begin{center}
	\includegraphics[width=0.7\linewidth]{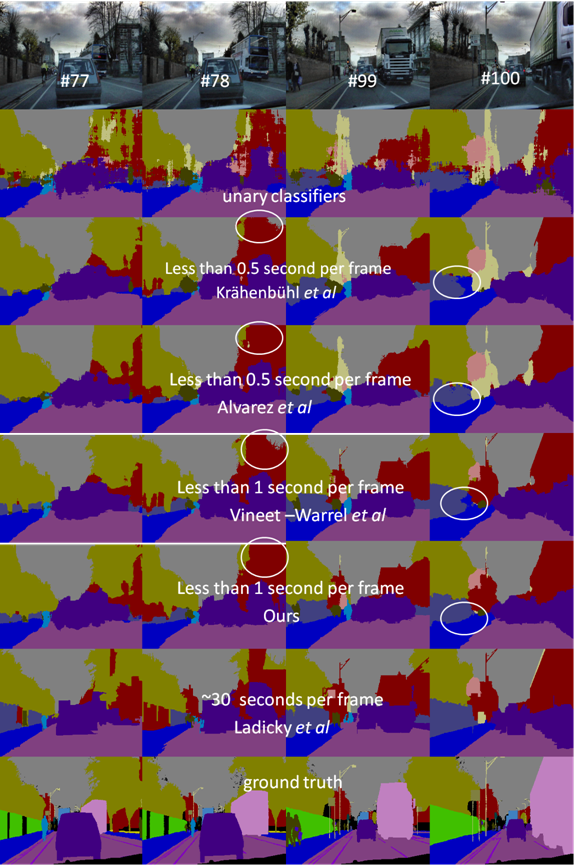}
\end{center}
\caption{
Qualitative results on Camvid dataset \cite{CamVid09}. From top to Bottom : Input frames, Unary potentials from TextonBoost classifier scores \cite{TextonBoost09}; Frame-level Dense-CRF \cite{DenseCRF13}; Video-Level Dense-CRF \cite{Alvarez14} shows improved temporal consistency over frame-level operation (previous row) without additional time overhead; frame-level Dense-CRF with $\operatorname{P^{n}-Potts}$ model \cite{vineet2014filter}; Proposed video-level Dense-CRF with $\operatorname{P^{n}-Potts}$ Model shows improved temporal consistency over the frame-level operation (previous row) without additional time-overhead; frame-level Graph-cut based slow inference with $\operatorname{P^{n}-Potts}$ Model and the Ground truth levels. }
\label{fig:Inference-Results}
%\vspace{2mm}
\end{figure*}

Figure \ref{fig:Inference-Results} shows some qualitative results of semantic segmentation in Camvid video dataset \cite{CamVid09}. In this particular experiment, we used the TextonBoost \cite{TextonBoost09} unary potentials for easy comparison with other recent methods. Video-Level Dense-CRF \cite{Alvarez14} shows improved temporal consistency over frame-level operation \cite{DenseCRF13} (previous row) without additional time overhead. 
For, pattern-based potentials, we use three different superpixel segmentations by varying parameters of the meanshift algorithm. Frame-level Dense-CRF with this $\operatorname{P^{n}-Potts}$ model \cite{vineet2014filter} almost achieves similar quality as of previous graph-cut based slow inference method \cite{Ladicky2009}, but lacks temporal consistency. The proposed 
video-level Dense-CRF with $\operatorname{P^{n}-Potts}$ model shows improved temporal consistency over the frame-level operation (previous row) without additional time-overhead. Video-Level dense CRF \cite{Alvarez14} and the proposed method perform inference on $50$ frames at once. On this dataset, with TextonBoost unaries our proposed method achieves $8$\% \ more accuracy than \cite{Alvarez14} by virtue of $\operatorname{P^{n}-Potts}$ model and $1.5$\% \ more accuracy over \cite{vineet2014filter} without additional time overhead by virtue of co-labeling.  
CNN feature classification yields improved unary potentials compared to the unaries provided by TextonBoost. Analyzing the final video semantic segmentation accuracy using CNN based unaries and proposed dense-CRF with $\operatorname{P^{n}-Potts}$ model remains our future work. 
%\newline
%\newline
%\newline

\bibliography{egbib}

\end{document}